\documentclass{article}
\usepackage{ijcai24}

\usepackage{times}
\usepackage{soul}
\usepackage{url}
\usepackage[hidelinks]{hyperref}
\usepackage[utf8]{inputenc}
\usepackage[small]{caption}
\usepackage{graphicx}
\usepackage{amsmath}
\usepackage{amsthm}
\usepackage{booktabs}
\usepackage{algorithm}
\usepackage{algorithmic}
\usepackage[switch]{lineno}

\usepackage{xspace}
\usepackage{soul}
\usepackage{url}
\usepackage{graphicx}
\usepackage{amsmath}
\usepackage{amsfonts}
\usepackage{multirow}
\usepackage{amsthm}
\usepackage{booktabs}
\usepackage{epstopdf}
\usepackage{xcolor}
\usepackage[labelformat=simple]{subcaption}
\usepackage{caption}
\usepackage{dsfont}
\usepackage{braket}
\usepackage{tablefootnote}
\usepackage{subcaption}
\usepackage{cleveref}
\usepackage{tcolorbox}

\newcommand{\ourmethod}{\texttt{ChatRule}\xspace}
\graphicspath{{images/}}

\title{\texttt{ChatRule}: Mining Logical Rules with Large Language Models for\\ Knowledge Graph Reasoning}

\author{
    Linhao Luo$^1$
\and
Jiaxin Ju$^2$\and
Bo Xiong$^{3}$\and
Yuan-Fang Li$^1$\and
Gholamreza Haffari$^1$\and
Shirui Pan$^2$\thanks{Corresponding author.}
\affiliations
$^1$Monash University,
$^2$Griffith University,
$^3$University of Stuttgart\\
\emails
\{linhao.luo, yuanfang.li, Gholamreza.Haffari\}@monash.edu,\\jiaxin.ju@griffithuni.edu.au, bo.xiong@ipvs.uni-stuttgart.de, s.pan@griffith.edu.au
}

\begin{document}
\maketitle
\begin{abstract}
Logical rules are essential for uncovering the logical connections between relations, which could improve reasoning performance and provide interpretable results on knowledge graphs (KGs). Although there have been many efforts to mine meaningful logical rules over KGs, existing methods suffer from computationally intensive searches over the rule space and a lack of scalability for large-scale KGs. Besides, they often ignore the semantics of relations which is crucial for uncovering logical connections. Recently, large language models (LLMs) have shown impressive performance in the field of natural language processing and various applications, owing to their emergent ability and generalizability. In this paper, we propose a novel framework, \ourmethod, unleashing the power of large language models for mining logical rules over knowledge graphs. Specifically, the framework is initiated with an LLM-based rule generator, leveraging both the semantic and structural information of KGs to prompt LLMs to generate logical rules. To refine the generated rules, a rule ranking module estimates the rule quality by incorporating facts from existing KGs. Last, the ranked rules can be used to conduct reasoning over KGs. \ourmethod is evaluated on four large-scale KGs, w.r.t. different rule quality metrics and downstream tasks, showing the effectiveness and scalability of our method\footnote{Code is available at: \url{https://github.com/RManLuo/ChatRule}}.
\end{abstract}

\section{Introduction}
\begin{figure}
    \includegraphics[width=\columnwidth]{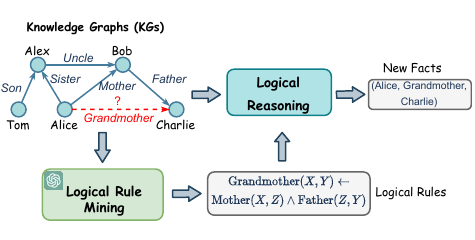}
    \caption{Illustration of mining logical rules for knowledge graphs reasoning with LLMs.}\label{fig:intro}
\end{figure}

Knowledge graphs (KGs) store enormous real-world knowledge in a structural format of triples. KG reasoning, which aims to infer new knowledge from existing facts, is a fundamental task in KGs and essential for many applications, such as KG completion \cite{qu2019probabilistic}, question-answering \cite{atif2023beamqa}, and recommendation \cite{wang2019explainable}. Recently, there has been an increasing need for interpretable KG reasoning, which can help users understand the reasoning process and improve the trustworthiness in high-stake scenarios, such as medical diagnosis \cite{liu2021neural} and legal judgment \cite{zhong2020iteratively}. Therefore, logical rules \cite{barwise1977introduction}, which are human-readable and can be generalized to different tasks, have been widely adopted for KG reasoning \cite{hou2021rule,liu2022tlogic}. For example, as shown in \Cref{fig:intro}, we can identify a logical rule: $\texttt{GrandMother}(X,Y)\leftarrow\texttt{Mother}(X,Z)\wedge\texttt{Father}(Z,Y)$ to predict missing facts for relation ``GrandMother''. 
To automatically discover meaningful rules from KGs for reasoning, logical rule mining has gained significant attention in the research community \cite{yang2017differentiable,sadeghian2019drum}.

Earlier studies on logical rule mining usually find logical rules by discovering the co-occurrences of frequent patterns in KG structure \cite{galarraga2013amie,chen2016ontological}. However, they usually require the enumeration of all possible rules on KGs and ranking them by estimated importance \cite{lao2010relational}. Although recent research has proposed to use deep learning methods to rank the rules. They are still limited by the exhaustive enumeration of rules and cannot scale to large-scale KGs \cite{yang2017differentiable,sadeghian2019drum}. 

Some recent methods address this issue by sampling paths from KGs and training models on them to capture the logical connections that form rules \cite{qu2020rnnlogic,cheng2022rlogic,cheng2022neural}. But they usually ignore contributions of relation semantics for expressing logical connections. For example, in commonsense, we know the ``mother'' of a person's ``father'' is his ``grandmother''. Based on this, we can define a rule like $\texttt{GrandMother}(X, Y) \leftarrow \texttt{Mother}(X,Z) \wedge \texttt{Father}(Z, Y)$ to express the logical connection. Whereas, due to the number of relations in KGs, it could be burdensome to ask domain-experts to annotate rules for each relation. Therefore, it is essential to automatically incorporate both the structure and semantics of relations to discover logical rules in KGs.

Large language models (LLMs) such as ChatGPT\footnote{\url{https://openai.com/blog/chatgpt}} and BARD\footnote{\url{https://bard.google.com/}} exhibit great ability in understanding natural language and handling many complex tasks \cite{zhao2023survey}. Trained on large-scale corpora, LLMs store a great amount of commonsense knowledge that can be used to facilitate KG reasoning \cite{pan2023unifying}. At the same time, LLMs are not designed to understand the structure of KGs, making it difficult to directly apply them to mining logical rules over KGs. Moreover, the widely acknowledged hallucination problem could make LLMs generate meaningless logical rules \cite{ji2023survey}.

To mitigate the gap between LLMs and logical rule mining, we propose a novel framework called \ourmethod, which leverages both the semantic and structural information of KGs to prompt LLMs to generate logical rules. Specifically, we first present an LLM-based \emph{rule generator} to generate candidate rules for each relation. We sample some paths from KGs to represent the structural information, which are then used in a carefully designed prompt to leverage the capabilities of LLMs for rule mining. To reduce the hallucination problem, we design a \emph{logical rule ranker} to evaluate the quality of generated rules and filter out meaningless rules by encompassing the observed facts in KGs. The quality scores are further used in the logical reasoning stage to reduce the impact of low-quality rules. Lastly, we feed the ranked rules into a \emph{logical reasoning} module to conduct interpretable reasoning over KGs. In our framework, the mined rules can be directly used for downstream tasks without any model training. Extensive experiments on four large-scale KGs demonstrate that \ourmethod significantly outperforms state-of-the-art methods on both knowledge graph completion and rule quality evaluation.

The main contributions of this paper are summarized as follows:
\begin{itemize}
    \item We propose a framework called \ourmethod that leverages the advantage of LLMs for mining logical rules. To the best of our knowledge, this is the first work that applies LLMs for logical rule mining.
    \item We present an end-to-end pipeline that utilizes the reasoning ability of LLMs and structure information of KGs for rule generation, rule ranking, and rule-based logical reasoning.
    \item We conduct extensive experiments on four datasets. Experiment results show that \ourmethod significantly outperforms state-of-the-art methods.
\end{itemize}
\section{Related Works}

\subsection{Logical Rule Mining}
Logical rule mining, which focuses on extracting meaningful rules from KGs, has been studied for a long time. Traditional methods enumerate the candidate rules, then access the quality of them by calculating weight scores \cite{lao2010relational,galarraga2013amie}. With the advancement of deep learning, researchers explore the idea of simultaneously learning logic rules and weights in a differentiable manner \cite{yang2017differentiable,sadeghian2019drum,yang2019learn}. However, these methods still conduct heavy optimization on the rule space, which limits their scalability. Recently, researchers have proposed to sample paths from KGs and train models on them to learn the logical connections. RLvLR \cite{omran2018scalable} samples rules from a subgraph and proposes an embedding-based score function to estimate the importance of each rule. RNNLogic \cite{qu2020rnnlogic} separates the rule generation and rule weighting, which can mutually enhance each other and reduce the search space. R5 \cite{lu2021r5} proposes a reinforcement learning framework that heuristically searches over KGs and mines underlying logical rules. NCRL \cite{cheng2022neural} predicts the best composition of rule bodies to discover rules. Ruleformer \cite{xu2022ruleformer} adopts a transformer-based model to encode context information and generate rules for the reasoning tasks, which is the state-of-the-art method in this area. Nevertheless, existing methods do not consider the semantics of relations and could lead to a suboptimal result. 

\subsection{Large Language Models}
Large language models (LLMs) are revolutionizing the field of natural language processing and artificial intelligence. Many LLMs (e.g., ChatGPT\footnotemark[2], Bard\footnotemark[3], FLAN \cite{wei2021finetuned}, and LLaMA \cite{touvron2023llama}) have demonstrated strong ability in various tasks.
Recently, researchers have also explored the possibility of applying LLMs to address KG tasks \cite{pan2023unifying,luo2023reasoning}. To better access the potential of LLMs, researchers design some prompts with few-shot examples \cite{brown2020language} or chain-of-thought reasoning \cite{wei2022chain} to maximize their ability. Nevertheless, these methods are not designed for logical rule mining, which requires the LLMs to understand both the structure of KGs and semantics of relations for generating meaningful rules.
\begin{figure*}
    \centering
    \includegraphics[width=.8\linewidth]{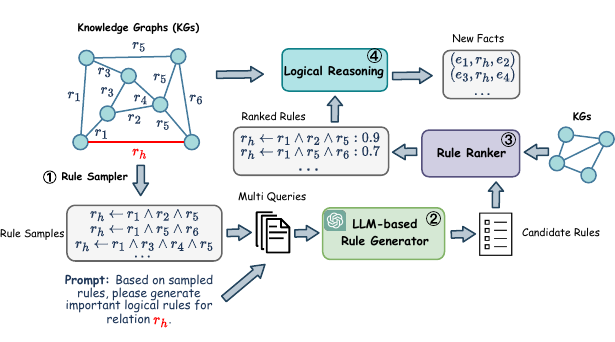}
    \caption{The overall framework of \ourmethod. 1) we first sample a few rule instances from the knowledge graph for a given target relation $r_h$. 2) we prompt the large language model (e.g., ChatGPT) to generate a set of coarse candidate rules. 3) we propose a rule ranker to estimable the quality of the generated rules based on facts in KGs. 4) the final rules can be applied for logical reasoning and addressing downstream tasks, such as knowledge graph completion.}\label{fig:framework}
\end{figure*}
\section{Preliminary and Problem Definition}

\noindent\textbf{Knowledge Graphs (KGs)} represent collections of facts in a form of triples $\mathcal{G}=\{(e,r,e')\subseteq \mathcal{E}\times \mathcal{R} \times \mathcal{E}\}$, where $e, e'\in\mathcal{E}$ and $r\in\mathcal{R}$ respectively denote the set of entities and relations.

\noindent\textbf{Logical Rules} are special cases of first-order logic \cite{barwise1977introduction}, which could facilitate interpretable reasoning on KGs \cite{yang2017differentiable}. Logical rules $\rho$ state the logical implication in the following form
\begin{equation}
    \label{eq:rule}
    \rho := r_h(X,Y) \leftarrow r_1(X, Z_1) \wedge \cdots \wedge r_L(Z_{L-1}, Y),
\end{equation}
where $\texttt{body}(\rho) := r_1(X, Z_1) \wedge \cdots \wedge r_L(Z_{L-1}, Y)$ denotes the conjunction of a series of relations called \textit{rule body}, $r_h(X,Y)$ denotes the \textit{rule head}, and $L$ denotes the length of rules. If the conditions on the rule body are satisfied, then the statement on the rule head also holds. 

An instance of the rule is realized by replacing the variables $X,Y,Z_*$ with actual entities in KGs. For example, given a rule $\texttt{GrandMother}(X, Y) \leftarrow \texttt{Mother}(X,Z_1) \wedge \texttt{Father}(Z_1, Y)$, one rule instance $\delta$ could be 
\begin{equation}
    \begin{split}
        &\texttt{GrandMother}(\text{Alice}, \text{Charlie})\leftarrow \\
        &\texttt{Mother}(\text{Alice}, \text{Bob}) 
        \wedge \texttt{Father}(\text{Bob}, \text{Charlie}),
    \end{split}\label{eq:rule_instance}
\end{equation}
which means that if Alice is the mother of Bob and Bob is the father of Charlie, then Alice is the grandmother of Charlie. 

\noindent\textbf{Problem Definition.} Given a target relation $r_h\in\mathcal{R}$ as the rule head, the goal of logical rule mining is to find a set of meaningful rules $P_{r_h} =\{\rho_1,\cdots,\rho_K\}$ that capture the logical connections of other relations to express the target relation $r_h$ in KGs.

\section{Approach}

In this section, we will introduce the proposed framework, called \ourmethod, for mining logical rules over KGs with large language models. The overall framework is illustrated in \Cref{fig:framework}, which contains three main components: 1) an LLM-based rule generator that leverages both the semantics and structural information to generate meaningful rules. 2) a rule ranker to estimate the quality of generated rules on KGs, and 3) a rule-based logical reasoning module to conduct reasoning over KGs for downstream tasks.

\subsection{LLM-based Rule Generator}
Conventional studies on logical rule mining usually focus on using structural information, \cite{galarraga2013amie,cheng2022neural}, which ignores the contributions of relation semantics for expressing logical connections. To harness the semantics understanding ability of large language models (LLMs), we propose an LLM-based rule generator that leverages both the semantics and structural information of KGs in generating meaningful rules.


\subsubsection{Rule Sampler}
To enable LLMs to understand KG structures for rule mining, we adopt a breadth-first search (BFS) sampler to sample a few \emph{closed-paths} from KGs, which can be treated as the instances of logical rule \cite{omran2018scalable,cheng2022rlogic}. Given a triple $(e_1,r_h,e_L)$, the closed-path is defined as a sequence of relations $r_1,\cdots,r_{L}$ that connects $e_1$ and $e_L$ in KGs, i.e., $e_1 \xrightarrow{r_1} e_2 \xrightarrow{r_2} \cdots \xrightarrow{r_{L}} e_L$. For example, given a triple $(\text{Alice}, \text{GrandMother}, \text{Charlie})$, a closed-path $p$ can be found as 
\begin{equation}
    p := \text{Alice} \xrightarrow{\texttt{Mother}} \text{Bob} \xrightarrow{\texttt{Father}} \text{Charlie}, 
\end{equation}
which closes the triple $(\text{Alice}, \text{GrandMother}, \text{Charlie})$ in KGs. By treating the triple as the rule head and closed-path as the rule body, we can obtain the rule instance $\delta$ shown in \Cref{eq:rule_instance}.

Given a target relation $r_h$, we first select a set of seed triples $\{(e, r_h, e')\}$ from KGs, from which we conduct BFS to sample a set of closed-paths $\{p\}$ with lengths less than $L$ to constitute a set of rule instances $\{\delta\}$. Following that, we substitute the actual entities in the rule instances with variables to obtain the rule samples $S_{r_h}=\{\rho\}$. The rule samples convey the structural information of KGs in a sequential format, which can be fed into the large language model to facilitate rule generation. 

\subsubsection{LLM-based Rule Generation}
Large language models (LLMs) trained on large-scale corpora exhibit the ability to understand the semantics of natural language and perform complex reasoning with commonsense knowledge \cite{zhou2020evaluating,tan2023evaluation}. To incorporate the structure and semantics information, we design a meticulously crafted prompt to harness the ability of LLMs for rule mining. 

For each rule from $S_{r_h}$ obtained by the rule sampler for a target relation $r_h$, we verbalize it into a natural-language sentence by removing the special symbols in the relation names, which could deteriorate the semantic understanding of LLMs. For the inverse of an original relation (i.e., $\texttt{wife}^{-1}$), we verbalize it by adding an ``inv\_'' symbol. Then, we place the verbalized rule samples into the prompt template and feed them into the LLMs (e.g., ChatGPT) to generate the rules. An example of the rule generation prompt and results of LLMs for relation ``\texttt{husband}(X,Y)'' is shown in \Cref{fig:rule_gen_prompt}. The detailed prompt can be found in the \textit{Appendix}.

\begin{figure}
    \centering
    \includegraphics[width=.97\columnwidth]{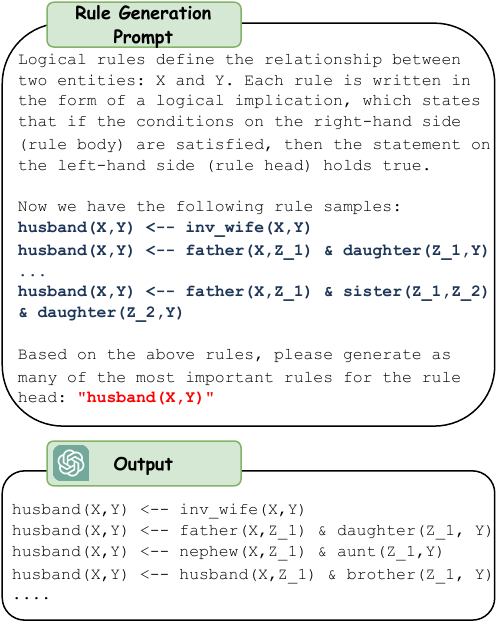}
    \caption{An example of the rule generation prompt and results of LLMs for relation ``\texttt{husband}(X,Y)''.}\label{fig:rule_gen_prompt}
\end{figure}

\subsubsection{Multi-Queries Rule Generation}
Due to the large number of rule samples, they cannot all be fed into the LLMs at once due to exceeding the context limitation. Thus, we divide the rule samples into $d$ different queries. Each query contains $k$ rule samples that are randomly selected from $S_{r_h}$. Then we prompt LLMs with queries separately and gather the responses of LLMs to obtain a set of candidate rules $C_{r_h}=\{\rho\}$.

\subsection{Logical Rule Ranking}\label{sec:ranking}
LLMs are known for having hallucination issues, which could generate incorrect results \cite{ji2023survey}. For example, the generated rule $\texttt{husband}(X,Y) \leftarrow \texttt{husband}(X,Z\_1)~\&~\texttt{brother}(Z\_1, Y)$, shown in the results of \Cref{fig:rule_gen_prompt}, is incorrect. Therefore, we develop a rule ranker to detect the hallucination and estimate the quality of generated rules based on facts in KGs.

The rule ranker aims to assign a quality score $s(\rho)$ for each rule $\rho$ in the candidate rule set $C_{r_h}$. Motivated by previous rule mining works \cite{galarraga2013amie}, we employ four measures, namely support, coverage, confidence, and PCA confidence, to assess the quality of rules. A detailed introduction and examples of each measure can be found in the \textit{Appendix}.

\noindent\textbf{Support} denotes the number of facts in KGs that satisfy the rule $\rho$, which is defined as
\begin{equation}
    \label{eq:support}
    \resizebox{.9\columnwidth}{!}{$
    \begin{split}
    \text{supp}(\rho) :=~&\#(e,e'): \exists (e,r_1,e_2)\wedge, \cdots, \wedge (e_{L-1}, r_L, e'): \\ &\texttt{body}(\rho) \wedge (e,r_h,e')\in\mathcal{G},
    \end{split}
    $}
\end{equation}
where $(e_1,r_1,e_2), \cdots, (e_{L-1}, r_L, e')$ denotes a series of facts in KGs that satisfy rule body $\texttt{body}(\rho)$ and $(e,r_h,e')$ denotes the fact satisfying the rule head $r_h$. 

Clearly, the rules that have zero support can be easily pruned away from the candidate set without any further refinement. However, support is an absolute number that could be higher for relations with more facts in KGs and provide biased ranking results.


\noindent\textbf{Coverage} normalizes the support by the number of facts for each relation in KGs, which is defined as
\begin{equation}
    \text{cove}(\rho) := \frac{\text{supp}(\rho)}{\#(e,e'): (e,r_h,e')\in\mathcal{G}}.
\end{equation}

The coverage quantifies the ratio of the existing facts in KGs that are implied by the rule $\rho$. To further consider the incorrect predictions of the rules, we introduce the confidence and PCA confidence to estimate the quality of rules.

\noindent\textbf{Confidence} is defined as the ratio of the number of facts that satisfy the rule $\rho$ and the number of times the rule body $body(\rho)$ is satisfied in KGs, which is defined as
\begin{equation}
    \text{conf}(\rho) := \frac{\text{supp}(\rho)}{\#(e,e'): \texttt{body}(\rho)\in\mathcal{G}}.
\end{equation}

The confidence assumes that all the facts derived from the rule body should be contained in KGs. However, the KGs are often incomplete in practice, which could lead to the missing of evidence facts. Therefore, we introduce the PCA confidence to select rules that could better generalize to unseen facts.

\noindent\textbf{PCA Confidence} is based on the theory of partial completeness assumption (PCA) \cite{galarraga2013amie} which is defined as the ratio of the number of facts that satisfy the rule $\rho$ and the number of times the rule body $\texttt{body}(\rho)$ is satisfied in the partial completion of KGs, which is defined as
\begin{gather}
    \begin{split}
        \text{partial}(\rho)(e,e') := ~&(e,r_1,e_2)\wedge, \cdots, \wedge (e_{L-1}, r_L, \hat{e}): \\ &\texttt{body}(\rho) \wedge (e,r_h, \hat{e}),
    \end{split}\\
    \text{pca}(\rho) := \frac{\text{supp}(\rho)}{\#(e,e'): \text{partial}(\rho)(e,e')\in\mathcal{G}}.
\end{gather}

The denominator of PCA confidence is not the size of the entire set of facts derived from the rule body. Instead, it is based on the number of facts that we know to be true along with those that we assume to be false. Therefore, the PCA confidence is better for estimating the quality and generalizability of rules in incomplete KGs. Experiment results in rule quality evaluation also support this claim.





\subsection{Rule-based Logical Reasoning}\label{sec:reasoning}
After logical rule ranking, we obtain a set of ranked rules $R_{r_h}=\{(\rho,s(\rho))\}$ for the target relation $r_h$. The ranked rules can be used for logical reasoning and addressing downstream tasks, such as knowledge graph completion, by applying existing algorithms such as forward chaining \cite{salvat1996sound}. 

Given a query $(e, r_h, ?)$, let $A$ be the set of candidate answers. For each $e' \in A$, we can apply the rule in $P_{r_h}$ to obtain the score as
\begin{equation}
    \text{score}(e') = \sum_{\rho\in R_{r_h}} \sum_{\texttt{body}(\rho)(e,e') \in \mathcal{G}} s(\rho),
\end{equation}
where $\texttt{body}(\rho)(e,e')$ denotes the path in the KGs that satisfies the rule body, and $s(\rho)$ denotes the quality score of the rule, which could be either converge, confidence, and PCA confidence. Then, we can rank the candidate answers $A$ based on the scores and select the top-$N$ answers as the final results.

\section{Experiment}

\subsection{Datasets}
In the experiment, we select four widely used datasets following previous studies \cite{cheng2022rlogic}: Family \cite{hinton1986learning}, UMLS \cite{kok2007statistical}, WN18RR \cite{dettmers2018convolutional}, and YAGO3-10 \cite{suchanek2007yago}. The statistics of the datasets are summarized in Table \ref{tab:dataset}.

\subsection{Baselines}
We compare our method against the SOTA rule mining baselines: AIME \cite{galarraga2013amie}, NeuralLP \cite{yang2017differentiable}, RNNLogic \cite{qu2020rnnlogic}, NLIL \cite{yang2019learn}, NCRL \cite{cheng2022neural}, and Ruleformer \cite{xu2022ruleformer}, on both knowledge graph completion and rule quality evaluation tasks.

\subsection{Metrics}
For the knowledge graph completion task, we mask the tail or head entity of each test triple and use the rule generated by each method to predict it. Following previous studies \cite{cheng2022neural}, we use the mean reciprocal rank (MRR) and the Hits@$N$ as the evaluation metrics and set $N$ to 1 and 10. For the rule quality evaluation task, we use the measures (e.g., support, coverage, confidence, and PCA confidence) discussed in the previous section on rule ranking.

\subsection{Experiment Settings}
For \ourmethod, we use the ChatGPT\footnote{We use the snapshot of ChatGPT taken from June 13th 2023 (gpt-3.5-turbo-0613) to ensure the reproducibility.} as the LLMs for rule generation. We select the PCA confidence as the final rule ranking measure and set maximum rule lengths $L$ to 3.
In the knowledge graph completion task, we follow the same settings as previous studies \cite{cheng2022rlogic,cheng2022neural}. For baselines, we use the publicized official implantation to conduct experiments. Detailed discussions about the settings can be found in the \textit{Appendix}.

\begin{table}[]
    \centering
    \caption{Dataset statistics}
    \label{tab:dataset}
        \begin{tabular}{@{}cccc@{}}
            \toprule
            Dataset  & \#Triples & \#Relation & \#Entity \\ \midrule
            Family   & 28,356    & 12         & 3,007    \\
            UMLS     & 5,960     & 46         & 135      \\
            WN18RR   & 93,003    & 11         & 40,943   \\
            YAGO3-10 & 1,089,040 & 37         & 123,182  \\ \bottomrule
        \end{tabular}%
\end{table}

\subsection{Knowledge Graph Completion}

Knowledge graph completion is a classical task that aims to predict the missing facts by using rule-based logical reasoning. This task has been adopted by various existing rule mining methods such as Neural-LP \cite{yang2017differentiable}, and NCRL \cite{cheng2022neural} to evaluate the quality of generated rules. We adopt rules generated by each method and use the same rule-based logical reasoning presented in \Cref{sec:reasoning} to predict the missing facts. The results are shown in \Cref{tab:kgc}.

From the results, we can observe that \ourmethod outperforms the baselines on most datasets. Specifically, the traditional method AIME, which only utilizes the structure information with inductive logic programming, has already achieved relatively good performance. However, AIME fails in large-scale KGs (e.g., YAGO3-10) due to the increasing number of relations and triples. Recent deep learning-based methods (e.g., Neural-LP, RNNLogic, and NLIL) achieve better performance by utilizing the learning ability of neural networks as they optimize the models on rule learning tasks. However, they suffer from the out-of-memory issue in handling large KGs due to the extensive rule-searching space. While NCRL samples close-paths to reduce the search space, it still ignores the semantics of relations, which leads to a suboptimal performance. Similar to the architecture of LLMs, Ruleformer adopts a transformer-based model to aggregate the context information from KGs and generate rules in a sequence-to-sequence manner, which achieves the second-best performance. This also demonstrates the great potential of the transformer architecture in logical rule mining. With the help of powerful pre-trained LLMs, \ourmethod can generate high-quality rules by incorporating the structure and semantic information of KGs. \ourmethod also sets new STOA performance in most settings.

\begin{table*}[]
    \centering
    \caption{Knowledge graph completion results. OOM denotes out-of-memory.}
    \label{tab:kgc}
    \resizebox{1\textwidth}{!}{%
        \begin{tabular}{@{}c|ccc|ccc|ccc|ccc@{}}
            \toprule
            \multirow{2}{*}{Methods} & \multicolumn{3}{c|}{Family} & \multicolumn{3}{c|}{UMLS} & \multicolumn{3}{c|}{WN18RR} & \multicolumn{3}{c}{YAGO3-10}                                                                                                                                         \\ \cmidrule(l){2-13}
                                     & MRR                         & Hits@1                    & Hits@10                     & MRR                          & Hits@1         & Hits@10        & MRR            & Hits@1         & Hits@10        & MRR            & Hits@1         & Hits@10        \\ \midrule
            AMIE                     & 0.778                       & 0.683                     & 0.891                       & 0.312                        & 0.195          & 0.560          & 0.162          & 0.060          & 0.278          & 0.012          & 0.008          & 0.021          \\
            Neural-LP                & 0.785                       & 0.720                     & 0.863                       & 0.505                        & 0.415          & 0.638          & 0.228          & 0.223          & 0.235          & OOM            & OOM            & OOM            \\
            RNNLogic                 & 0.860                       & 0.792                     & 0.957                       & 0.750                        & 0.630          & 0.924          & 0.216          & 0.183          & 0.275          & OOM            & OOM            & OOM            \\
            NLIL                     & 0.358                       & 0.321                     & 0.416                       & 0.693                        & 0.632          & 0.921          & 0.223          & 0.222          & 0.225          & OOM            & OOM            & OOM            \\
            NCRL                     & 0.826                       & 0.725                     & 0.963                       & 0.728                        & 0.576          & 0.938          & 0.316          & 0.272          & 0.397          & 0.234          & 0.181          & 0.334          \\
            Ruleformer               & 0.897                       & 0.841                     & 0.963                       & 0.691                        & 0.555          & 0.863          & 0.292          & 0.258          & 0.355          & \textbf{0.527} & \textbf{0.520} & 0.535          \\ \midrule
            \ourmethod (ChatGPT)     & \textbf{0.906}              & \textbf{0.854}            & \textbf{0.968}              & \textbf{0.780}               & \textbf{0.685} & \textbf{0.948} & \textbf{0.335} & \textbf{0.301} & \textbf{0.400} & 0.449          & 0.354          & \textbf{0.627} \\ \bottomrule
        \end{tabular}%
    }
\end{table*}

\begin{table*}[]
    \centering
    \caption{Rule quality evaluation results.}
    \label{tab:rule_quality}
    \resizebox{1\textwidth}{!}{%
        \begin{tabular}{@{}c|cccc|cccc@{}}
            \toprule
            \multirow{2}{*}{Methods} & \multicolumn{4}{c|}{Family} & \multicolumn{4}{c}{UMLS}                                                                                                                                \\ \cmidrule(l){2-9}
                                     & Support                     & Coverage                     & Confidence    & PCA confidence & Support          & Coverage      & Confidence    & PCA confidence \\ \midrule
            AMIE                     & 243.90                      & 0.11                         & 0.17          & 0.30           & 35.52             & 0.10          & 0.13          & 0.21           \\
            Neural-LP                & 280.94                      & 0.13                         & 0.21          & 0.30           & 11.59             & 0.06          & 0.08          & 0.09           \\
            NCRL                     & 179.96                      & 0.09                         & 0.12          & 0.16           & 13.25             & 0.03          & 0.04          & 0.06           \\
            Ruleformer               & 325.32                      & 0.15                         & 0.22          & 0.32           & \textbf{57.00}    & 0.25          & \textbf{0.33} & 0.37           \\
            \ourmethod (ChatGPT)     & \textbf{403.04}             & \textbf{0.17}                & \textbf{0.23} & \textbf{0.34}  & 28.06             & \textbf{0.32} & 0.21          & \textbf{0.38}  \\ \midrule\midrule
            \multirow{2}{*}{Methods} & \multicolumn{4}{c|}{WN18RR} & \multicolumn{4}{c}{YAGO3-10}                                                                                                                            \\ \cmidrule(l){2-9} 
                                     & Support                     & Coverage                     & Confidence    & PCA Confidence & Support          & Coverage      & Confidence    & PCA confidence \\ \midrule
            AMIE                     & 378.76                      & 0.07                         & 0.08          & 0.10           & 758.28            & 0.06          & 0.06          & 0.07           \\
            Neural-LP                & 285.24                      & 0.05                         & 0.06          & 0.09           & -                 & -             & -             & -              \\
            NCRL                     & 400.38                      & 0.18                         & 0.02          & 0.03           & 12660.16          & 0.13          & 0.12          & 0.05           \\
            Ruleformer               & 325.32                      & 0.15                         & \textbf{0.22} & \textbf{0.32}  & 495.79            & 0.15          & \textbf{0.22} & \textbf{0.22}  \\
            \ourmethod (ChatGPT)     & \textbf{667.28}             & \textbf{0.12}                & 0.09          & 0.16           & \textbf{13656.41} & \textbf{0.27} & 0.09          & 0.14           \\ \bottomrule
        \end{tabular}%
    }
\end{table*}

\subsection{Rule Quality Evaluation}
To further demonstrate the effectiveness of the four measures (i.e., support, coverage, confidence, and PCA confidence) adopted in the rule ranking, we use them to evaluate the rules generated by each method. The results are shown in \Cref{tab:rule_quality}.

From the results, we can observe that \ourmethod can generate rules with higher support, coverage, and confidence than the baselines. Specifically, we can observe that the scores of the measures are consistent with the performance in knowledge graph completion. This demonstrates that the selected measures can well quantify the quality of rules. 
Notably, while NCRL achieves higher scores than Ruleformer in the support metric of YAGO3-10 (12660.16 vs 495.79), NCRL is still outperformed by Ruleformer in knowledge graph completion. This is because the rules generated by Ruleformer have a better PCA confidence, which is more suitable for evaluating rules in incomplete KGs. Similarly, a higher PCA confidence score indicates that \ourmethod can generate rules with better generalizability instead of solely relying on sampled rules from prompts. Consequently, \ourmethod also demonstrates superior performance in the task of knowledge graph completion.


\begin{table}[]
    \centering
    \caption{Performance of knowledge graph completion using rules generated by different LLMs on the Family dataset.}
    \label{tab:llms}
    \begin{tabular}{@{}lccc@{}}
        \toprule
        Model               & MRR            & Hits@1         & Hits@10        \\ \midrule
        ChatGPT             & \textbf{0.849} & \textbf{0.765} & \textbf{0.964} \\
        GPT-4               & 0.803          & 0.704          & 0.926          \\
        Mistral-7B-Instruct & 0.735          & 0.652          & 0.912          \\
        Llama2-chat-7B      & 0.742          & 0.636          & 0.893          \\
        Llama2-chat-7B-13B  & 0.767          & 0.671          & 0.900          \\
        Llama2-chat-7B-70B  & 0.731          & 0.626          & 0.875          \\ \bottomrule
    \end{tabular}%
\end{table}

\subsection{Ablation Studies}

\noindent\textbf{Analysis of different LLMs.} We first evaluate the performance of \ourmethod with several LLMs at different sizes, including GPT-4 \cite{openai2023gpt4}, Mistral-7B-Instruct \cite{jiang2023mistral}, and LLaMA2-Chat-7B/13B/70B \cite{touvron2023llama}. The details of model versions are available in \textit{Appendix}.

From the results shown in \Cref{tab:llms}, we can observe that \ourmethod with different LLMs still achieves a comparable performance against baselines. This demonstrates the generalizability of \ourmethod. An interesting finding is that the performance of \ourmethod is not always improved by using larger LLMs. For instance, \ourmethod performs better with ChatGPT compared to GPT-4. In the LLaMA2 family, LLaMA2-Chat-7B outperforms LLaMA2-Chat-70B but is surpassed by LLaMA2-Chat-13B. One reason for this is that different LLMs are sensitive to the input prompt; they may behave differently and achieve varying levels of performance despite having the same input prompt. Thus, further adjusting the input prompt for different LLMs may achieve better performance. Another reason is that larger LLMs tend to generate fewer rules compared to smaller ones. We hypothesize this could be due to the heavier computation cost of larger LLMs, which could lead to a higher probability of early stopping. The reduced number of rules may result in poorer coverage of ground-truth rules and suboptimal performance.



\begin{table}[]
    \centering
    \caption{Analysis of each ranking measure.}
    \label{tab:ranking}
    \resizebox{\columnwidth}{!}{%
        \begin{tabular}{@{}c|ccc|ccc@{}}
            \toprule
            \multirow{2}{*}{Ranking   Metrics} & \multicolumn{3}{c|}{Family} & \multicolumn{3}{c}{UMLS}                                                                     \\ \cmidrule(l){2-7}
                                               & MRR                         & Hits@1                   & Hits@10        & MRR            & Hits@1         & Hits@10        \\ \midrule
            None                               & 0.756                       & 0.630                    & 0.940          & 0.413          & 0.327          & 0.605          \\
            Coverage                           & 0.795                       & 0.673                    & 0.956          & 0.604          & 0.460          & 0.843          \\
            Confidence                         & 0.872                       & 0.799                    & 0.967          & 0.700          & 0.602          & 0.898          \\
            PCA   Confidence                   & \textbf{0.906}              & \textbf{0.854}           & \textbf{0.968} & \textbf{0.780} & \textbf{0.685} & \textbf{0.948} \\ \bottomrule
        \end{tabular}%
    }
\end{table}

\begin{figure}
    \includegraphics[width=.9\columnwidth]{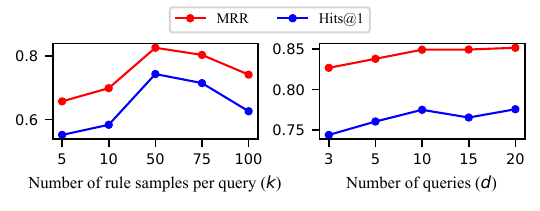}
    \caption{Parameter analysis of number of rule samples per query ($k$) and number of queries ($d$) on the Family dataset.}\label{fig:parameter}
\end{figure}

\noindent\textbf{Analysis of ranking measures.}
We test the effectiveness of each measure (i.e., coverage, confidence, and PCA confidence) adopted in rule ranking. The rules are all generated by ChatGPT on the Family and UMLS datasets.

The results are shown in \Cref{tab:ranking}. From the results, we can see that when one of the ranking measures is employed, the performance of \ourmethod is improved over when no ranking measure (i.e., None) is employed. This demonstrates that the ranking measure can effectively reduce the impact of low-quality rules. The PCA confidence metric achieves the best performance among all the ranking measures. This show that PCA confidence enables to quantify the quality of rules in incomplete KGs and selects rules with better generalizability, which is also chosen as the final ranking metrics.

\noindent\textbf{Analysis of hyperparameters.} Last, we evaluate the performance of \ourmethod with different hyperparameters, i.e., different numbers of rule samples per query ($k$) and different numbers of queries ($d$) on the Family dataset. The results are shown in \Cref{fig:parameter}.

From results, we can observe that the performance of \ourmethod first improves with the increase of $k$ and slightly drops once $k$ reaches 50. This shows that with more rule samples in the prompt, \ourmethod can generate rules with better quality. However, too large a $k$ value leads to a longer prompt. Existing LLMs are known to suffer from understanding long contexts \cite{liu2023lost}, which could lead to a suboptimal performance. In contrast, the performance of \ourmethod continually improves with the increase of $d$. This demonstrates that \ourmethod can generate better rules by ``seeing'' more rule samples in KGs. However, more queries also introduce higher computational costs, which could limit the scalability of \ourmethod. Therefore, we set $k$ to 50 and $d$ to 10 in the experiment.


\subsection{Case Studies}
We first show the statistics of the mined rule the corresponding overall API cost\footnote{The costs are calculated based on the API price defined by OpenAI (i.e., 0.001\$ and 0.002\$ per 1000 input and output tokens).} for each dataset in \Cref{tab:statistics}. From results, we can observe that \ourmethod can mine a substantial number of meaningful rules at a low API cost.

We present some representative logic rules mined from different datasets in \Cref{tab:case}. The results show that the rules generated by our method are both interpretable and of high quality. For instance, ``wife'' is intuitively the inverse relation of ``husband'', the rule $\texttt{husband} \leftarrow \texttt{inv\_wife}$ is successfully extracted by \ourmethod with the semantics of the relationship considered. Similarly, ``playsFor'' is the synonym of ``isAffiliatedTo'', which constitutes the rule $\texttt{playsFor} \leftarrow \texttt{isAffiliatedTo}$. 
Additionally, the rules also follow commonsense knowledge, exhibiting great interpretability. For example, the rule $\texttt{diagnoses} \leftarrow \texttt{analyzes} \wedge \texttt{causes}$ shows that to make a diagnosis, doctors usually need to analyze the patient's symptoms and find the cause of the disease.
Last, the generated rules also uncover some associative logical connections. The rule $\texttt{isPoliticianOf} \leftarrow \texttt{hasChild}\wedge\texttt{isPoliticianOf}$ indicates that children usually inherit their parents' political position, which is supported by the support and PCA scores. The full lists of mined rules are available in the \textit{supplementary file}.

\begin{table}[]
    \centering
    \caption{Statistics of the mined rules and overall API cost for each dataset.}
    \label{tab:statistics}
    \resizebox{.97\columnwidth}{!}{%
        \begin{tabular}{@{}cccc@{}}
            \toprule
            Dataset & Total Rules & Avg. Rules per Relation & Cost (\$) \\ \midrule
            Family  & 1200        & 100                     & 0.880     \\
            UMLS    & 610        & 29                      & 2.623     \\
            WN18RR  & 451         & 41                      & 0.399     \\
            YAGO    & 1924        & 50                      & 1.409     \\ \bottomrule
        \end{tabular}%
    }
\end{table}


\begin{table}[]
    \centering
    \caption{Representative rules mined on each dataset.}
    \label{tab:case}
    \resizebox{\columnwidth}{!}{%
        \begin{tabular}{@{}c|ccc@{}}
            \toprule
            Datasets                  & Rule                                                                                      & Support & PCA Confidence \\ \midrule
            \multirow{3}{*}{Family}   & $\texttt{husband} \leftarrow   \texttt{inv\_wife} $                                       & 655     & 0.98           \\
                                      & $\texttt{father}   \leftarrow \texttt{husband} \wedge \texttt{mother} $                   & 1260    & 0.88           \\
                                      & $\texttt{aunt}   \leftarrow \texttt{sister} \wedge \texttt{aunt} $                        & 2152    & 0.79           \\ \midrule
            \multirow{3}{*}{UMLS}   
            & $\texttt{prevents} \leftarrow \texttt{complicates} \wedge \texttt{inv\_prevents} \wedge \texttt{causes}$ & 16 & 0.53 \\
            & $\texttt{treats} \leftarrow   \texttt{prevents} \wedge \texttt{inv\_treats} \wedge , \texttt{treats}$                                       & 51     & 0.92           \\
            & $\texttt{diagnoses} \leftarrow \texttt{analyzes} \wedge \texttt{causes}$ & 20 & 1.00 \\
            \midrule
            \multirow{3}{*}{WN18RR}   
            & $\texttt{\_has\_part} \leftarrow  \texttt{\_has\_part} \wedge \texttt{inv\_\_has\_part} \wedge \texttt{\_has\_part}$ & 4209 & 0.39
            \\
            & $\texttt{\_also\_see} \leftarrow \texttt{\_also\_see} \wedge \texttt{inv\_\_also\_see} \wedge \texttt{\_also\_see}$ & 1151 & 0.47
            \\ 
            & $\texttt{\_hypernym} \leftarrow \texttt{\_member\_meronym} \wedge \texttt{inv\_\_member\_meronym} \wedge \texttt{\_hypernym}$ & 1996 & 1.0 \\\midrule
            \multirow{3}{*}{YAGO3-10} & $\texttt{hasChild} \leftarrow   \texttt{isMarriedTo} \wedge \texttt{hasChild} $           & 1174    & 0.60           \\
                                      & $\texttt{isPoliticianOf   } \leftarrow \texttt{hasChild} \wedge \texttt{isPoliticianOf} $ & 720     & 0.86           \\
                                      & $\texttt{playsFor}   \leftarrow \texttt{isAffiliatedTo} $                                 & 229794  & 0.75           \\ \bottomrule
        \end{tabular}%
    }
\end{table}

\section{Conclusion}
In this paper, we introduce a new approach called \ourmethod for bridging the gap in logical rule mining on KGs. In \ourmethod, we propose a rule generator based on LLMs that incorporates both semantics and structural information to generate meaningful rules. Additionally, a rule ranker is developed to assess the quality of the rules and eliminate incorrect ones. Finally, the generated rules can be directly used for knowledge graph reasoning without additional model training. Extensive experiments on several datasets demonstrate \ourmethod can generate high-quality and interpretable rules for downstream tasks. In the future, we will explore integrating advanced models to enhance LLMs understanding of structural information and improve the performance of rule mining.

\bibliographystyle{named}
\bibliography{sections/main}
\appendix
\section{Ranking Measures}
In this section, we provide a detailed example of the ranking measures (i.e., support, coverage, confidence, and PCA confidence).
For example, given a rule $$\rho := \texttt{playsFor}   \leftarrow \texttt{isAffiliatedTo},$$ and a simple KG in \Cref{tab:exp_kgs}. The measures can be calculated as follows.

\noindent\textbf{Support} denotes the number of facts in KGs that satisfy the rule $\rho$, which is defined as
\begin{equation}
    \label{eq:support}
    \resizebox{.9\columnwidth}{!}{$
    \begin{split}
    \text{supp}(\rho) :=~&\#(e,e'): \exists (e,r_1,e_2)\wedge, \cdots, \wedge (e_{L-1}, r_L, e'): \\ &\texttt{body}(\rho) \wedge (e,r_h,e')\in\mathcal{G},
    \end{split}
    $}
\end{equation}
where $(e_1,r_1,e_2), \cdots, (e_{L-1}, r_L, e')$ denotes a series of facts in KGs that satisfy rule body $\texttt{body}(\rho)$ and $(e,r_h,e')$ denotes the fact satisfying the rule head $r_h$. 

\noindent\textbf{Example.} For KG given in \Cref{tab:exp_kgs}, support of the rule is 1, because (Alex, isAffiliatedTo, Club 1) and  (Alex, playsFor, Club 1) exist in KG.

\noindent\textbf{Coverage} normalizes the support by the number of facts for each relation in KGs, which is defined as
\begin{equation}
    \text{cove}(\rho) := \frac{\text{supp}(\rho)}{\#(e,e'): (e,r_h,e')\in\mathcal{G}}.
\end{equation}

\noindent\textbf{Example.} For KG given in \Cref{tab:exp_kgs}, coverage of the rule is 1/2, because we have 2 facts under the ``\texttt{playsFor}'' relation.

The coverage quantifies the ratio of the existing facts in KGs that are implied by the rule $\rho$. To further consider the incorrect predictions of the rules, we introduce the confidence and PCA confidence to estimate the quality of rules.

\noindent\textbf{Confidence} is defined as the ratio of the number of facts that satisfy the rule $\rho$ and the number of times the rule body $body(\rho)$ is satisfied in KGs, which is defined as
\begin{equation}
    \text{conf}(\rho) := \frac{\text{supp}(\rho)}{\#(e,e'): \texttt{body}(\rho)\in\mathcal{G}}.
\end{equation}
\noindent\textbf{Example.} For KG given in \Cref{tab:exp_kgs}, the confidence of the rule is 1/3, because there is one positive fact satisfy the rule, and there are two facts (i.e., (Alex, isAffiliatedTo, Club 2) and (Bob, isAffiliatedTo, Club 3)) are considered as negative examples.

The confidence assumes that all the facts derived from the rule body should be contained in KGs. However, the KGs are often incomplete in practice, which could lead to the missing of evidence facts. Therefore, we introduce the PCA (partial completeness assumption) confidence to select rules that could better generalize to unseen facts. PCA confidence only considers the hard negative examples, which contradict the facts in existing KGs, and PCA confidence ignores the soft negative examples, which we have zero knowledge of their correctness.

\noindent\textbf{PCA Confidence} is based on the theory of partial completeness assumption (PCA) \cite{galarraga2013amie} which is defined as the ratio of the number of facts that satisfy the rule $\rho$ and the number of times the rule body $\texttt{body}(\rho)$ is satisfied in the partial completion of KGs, which is defined as
\begin{gather}
    \begin{split}
        \text{partial}(\rho)(e,e') := ~&(e,r_1,e_2)\wedge, \cdots, \wedge (e_{L-1}, r_L, \hat{e}): \\ &\texttt{body}(\rho) \wedge (e,r_h, \hat{e}),
    \end{split}\\
    \text{pca}(\rho) := \frac{\text{supp}(\rho)}{\#(e,e'): \text{partial}(\rho)(e,e')\in\mathcal{G}}.
\end{gather}

\noindent\textbf{Example.} For KG given in \Cref{tab:exp_kgs}, the PCA confidence of the rule is 1/2, because (Alex, isAffiliatedTo, Club 2) is a hard negative example which violates the fact that (Alex, playsFor, Club 1), and (Bob, isAffiliatedTo, Club 3) is a soft negative example, which is removed from the denominator since we do not know whether ``Bob'' plays for ``Club 3'' or not, based on facts in existing KG.

\begin{table}[]
\centering
\caption{An example KG containing two relations and five facts}
\label{tab:exp_kgs}
\begin{tabular}{@{}ll@{}}
\toprule
\texttt{isAffiliatedTo} & \texttt{playsFor} \\ \midrule
(Alex, Club 1)     & (Alex, Club 1)           \\
(Alex, Club 2)     & (Charlie, Club 2)       \\
(Bob, Club 3)      &                         \\ \bottomrule
\end{tabular}%
\end{table}

\section{Datasets}
In the experiment, we select four widely used datasets following previous studies \cite{cheng2022rlogic}: Family \cite{hinton1986learning}, UMLS \cite{kok2007statistical}, WN18RR \cite{dettmers2018convolutional}, and YAGO3-10 \cite{suchanek2007yago}.
\begin{itemize}
    \item Family \cite{hinton1986learning} is a knowledge graph defines the relationships of members in a family, e.g., ``Father'', ``Mother'', and ``Aunt''.
    \item UMLS \cite{kok2007statistical} is a bio-medicine knowledge, where entities are biomedical concepts, and relations include treatments and diagnoses.
     \item WN18RR \cite{dettmers2018convolutional} is an English vocabulary knowledge graphs designed to organize words according to their semantic relationships. Words are connected by a series of relationships, including ``hypernym'', ``derivation'', etc.
    \item YAGO3-10 \cite{suchanek2007yago} is another large scale knowledge graph constructed from multiple data sources, like Wikipedia, WordNet, and GeoNames, which contains many relations, e.g., ``was born in'', ``lives in'', and ``politician of''.
\end{itemize}

\section{Baselines}
We compare our method against the SOTA rule mining baselines: AIME \cite{galarraga2013amie}, NeuralLP \cite{yang2017differentiable}, RNNLogic \cite{qu2020rnnlogic}, NLIL \cite{yang2019learn}, NCRL \cite{cheng2022neural}, and Ruleformer \cite{xu2022ruleformer} in experiments.

\begin{itemize}
    \item AIME\footnote{\url{https://github.com/dig-team/amie}} \cite{galarraga2013amie} is a conventional logical rule mining method, which discovers rules from KG with inductive logica programming.
    \item NeuralLP\footnote{\url{https://github.com/fanyangxyz/Neural-LP}} \cite{yang2017differentiable} proposes a neural logic programming, that learns logical rules in an end-to-end differential way.
    \item RNNLogic\footnote{\url{https://github.com/DeepGraphLearning/RNNLogic/tree/main}} \cite{qu2020rnnlogic} proposes a rule generator and a reasoning predictor with logic rules. It develop an EM-based algorithm for optimization and learning high-quality rules for reasoning.
    \item NLIL\footnote{\url{https://github.com/gblackout/NLIL}} \cite{yang2019learn} proposes a neural logic inductive learning model that is an efficient differentiable inductive logical programming framework that learns first-order logic rules.
    \item NCRL\footnote{\url{https://github.com/vivian1993/NCRL/commits/main}} \cite{cheng2022neural} infers rules by recurrently merging compositions in the rule body, which detects the best compositional structure of a rule body for expressing the rule head.
    \item Ruleformer\footnote{\url{https://github.com/zjukg/ruleformer}} \cite{xu2022ruleformer} is a transformer-based model that encodes the context information from KGs to generate rules.
\end{itemize}

\section{Large Language Models}
\label{appendix:llm}

The LLMs used in experiments are shown in \Cref{tab:llm}. For open-sourced LLMs, we utilize available checkpoints from HuggingFace\footnote{\url{https://huggingface.co/}}. For ChatGPT and GPT-4, we used the API provided by OpenAI\footnote{\url{https://api.openai.com/v1/chat/completions}}.

\begin{table}
    \centering
    \resizebox{\columnwidth}{!}{%
    \begin{tabular}{@{}ll@{}}
    \toprule
    \textbf{LLM}        & \textbf{Model Implementation}    \\
    \midrule
    Mistral-7B-Instruct& mistralai/Mistral-7B-Instruct-v0.1   \\
    LlaMA2-chat-7B    & meta-llama/Llama-2-7b-chat            \\
    LlaMA2-chat-13B   & meta-llama/Llama-2-13b-chat            \\
    LlaMA2 70B   & meta-llama/Llama-2-70b-chat            \\
    ChatGPT   & gpt-3.5-turbo-0613             \\
    GPT-4 & gpt-4-0613 \\
    \bottomrule
    \end{tabular}
    }
    \caption{Details of used LLMs.}
    \label{tab:llm}
\end{table}

\section{Experiment Settings}
For \ourmethod, we use the ChatGPT\footnote{We use the snapshot of ChatGPT taken from June 13th 2023 (gpt-3.5-turbo-0613) to ensure the reproducibility.} as the LLMs for rule generation. We select the PCA confidence as the final rule ranking measure and set maximum rule lengths $L$ to 3. In the knowledge graph completion task, we follow the same settings as previous studies \cite{cheng2022rlogic,cheng2022neural}. For baselines, we use the publicized codes for comparison. 


\section{Rule Generation Prompt}
The prompt template used for rule generation is shown as follows.

\begin{minipage}{.99\columnwidth}
    \centering
    \vspace{1mm}
    \begin{tcolorbox}[title=Rule Generation Prompt, fonttitle=\bfseries\small]
        \small
        Logical rules define the relationship between two entities: X and Y. Each rule is written in the form of a logical implication, which states that if the conditions on the right-hand side (rule body) are satisfied, then the statement on the left-hand side (rule head) holds true. \\\\
        Now we have the following rules: \\
        \textit{\{ rule samples \}} \\\\
        Based on the above rules, please generate as many of the most important rules for the rule head: ``\textit{\{rule head}\}(X,Y)'' as possible. Please only select predicates form: \textit{\{ relations in KGs \}}. Return the rules only without any explanations.
    \end{tcolorbox}
    \vspace{1mm}
  \end{minipage}


\end{document}